\documentclass[letterpaper, 10 pt, conference]{ieeeconf}
\IEEEoverridecommandlockouts
\usepackage{amsmath,amsfonts}

\usepackage{array}
\usepackage{csquotes}
\usepackage{textcomp}
\usepackage{stfloats}
\usepackage{url}
\usepackage{verbatim}

\usepackage{graphicx}
\usepackage{cite}
\usepackage{algpseudocode}
\usepackage{graphicx}
\usepackage{textcomp}
\usepackage{xcolor}
\usepackage{varwidth}
\usepackage{subfigure}
\usepackage{cite}
\usepackage{comment}
\usepackage{afterpage}
\usepackage{flushend}
\usepackage{stfloats}

\usepackage[algorithm,algosection]{algorithm}

\algrenewcommand\algorithmicrequire{\textbf{Initialize:}}

\makeatletter
\newcommand{\brokenline}[2][t]{\parbox[#1]{\dimexpr\linewidth-\ALG@thistlm}{\strut\raggedright #2\strut}}
\makeatother
\begin{document}

\title{
\LARGE \bf CONTHER: Context-Aware Reinforcement Learning for Robotic Manipulation with Sparse Rewards
}
\author{Maria Makarova$^{1}$, Qian Liu$^{2}$ and Dzmitry Tsetserukou$^{1}$
\thanks{$^{1}$The authors are with the Intelligent Space Robotics Laboratory, Center for Digital Engineering, Skolkovo Institute of Science and Technology (Skoltech), 121205 Moscow, Russia. 
{\tt \small $\{$maria.makarova2, d.tsetserukou$\}$@skoltech.ru}}
\thanks{$^{2}$ Qian Liu is with the Department of Computer
Science and Technology, Dalian University of Technology, China. {\tt\small qianliu@dlut.edu.cn}}
}
\maketitle
\begin{abstract}
%CONTHER: Human-Like Contextual Robot Learning via Hindsight Experience Replay and Transformers without Expert Demonstrations
This paper investigates whether sequential context improves goal-conditioned Reinforcement Learning in sparse-reward manipulation tasks. While Hindsight Experience Replay (HER) addresses reward sparsity through goal relabeling, its operation on isolated transitions limits its ability to capture temporal dependencies inherent in joint-space control. We hypothesize that incorporating motion history can enhance policy learning and introduce CONTHER, which integrates a Transformer-based architecture with a modified HER replay buffer. The Transformer encodes sequences of prior states and goals to provide temporal awareness, while the buffer populates experience with artificially successful trajectories. Two architectural variants are analyzed to examine how contextual information should be integrated. In simulated point-reaching tasks with a UR3 manipulator, CONTHER achieves a 38.46\% higher average success rate compared to baselines and outperforms the strongest baseline by 28.21\%, with faster convergence and more stable learning. The framework is further evaluated on three dynamic tasks requiring complex trajectory following and obstacle avoidance, where temporal context is critical. By operating directly on joint velocities, the approach provides a foundation for transfer to physical systems. The primary contribution is a systematic investigation into fusing sequential context with goal relabeling, offering insights into how temporal awareness benefits policy learning.
\end{abstract}

\section{Introduction}

Robotic manipulation, a field of considerable interest in the development of modern robotic systems, encompasses a broad array of techniques and methodologies. A recent trend in this area involves the implementation of Imitation Learning to address complex manipulation tasks. A prominent technique within this domain is Behavioral Cloning, along with its various modifications \cite{il1, il2,il3}. 
%The scalability of these algorithms necessitates large and high-quality demonstration datasets, typically obtained from human experts, to generalize information to novel scenarios\cite{demo1,demo2,demos}.
However, these methods require large expert demonstration datasets\cite{demo1,demos}.

%Another approach is building Reinforcement Learning (RL) models. RL algorithms encounter challenges in optimizing sparse reward functions due to the potential complexity of attaining goal states, which can be quite difficult, if not impossible, during the exploratory phase, which is typical for manipulation tasks. 

Reinforcement Learning (RL) offers an alternative approach but faces challenges with sparse rewards, as attaining goal states during exploration is often difficult in manipulation tasks.

\begin{figure}[h]
  \centering
  \includegraphics[width=0.55\linewidth]{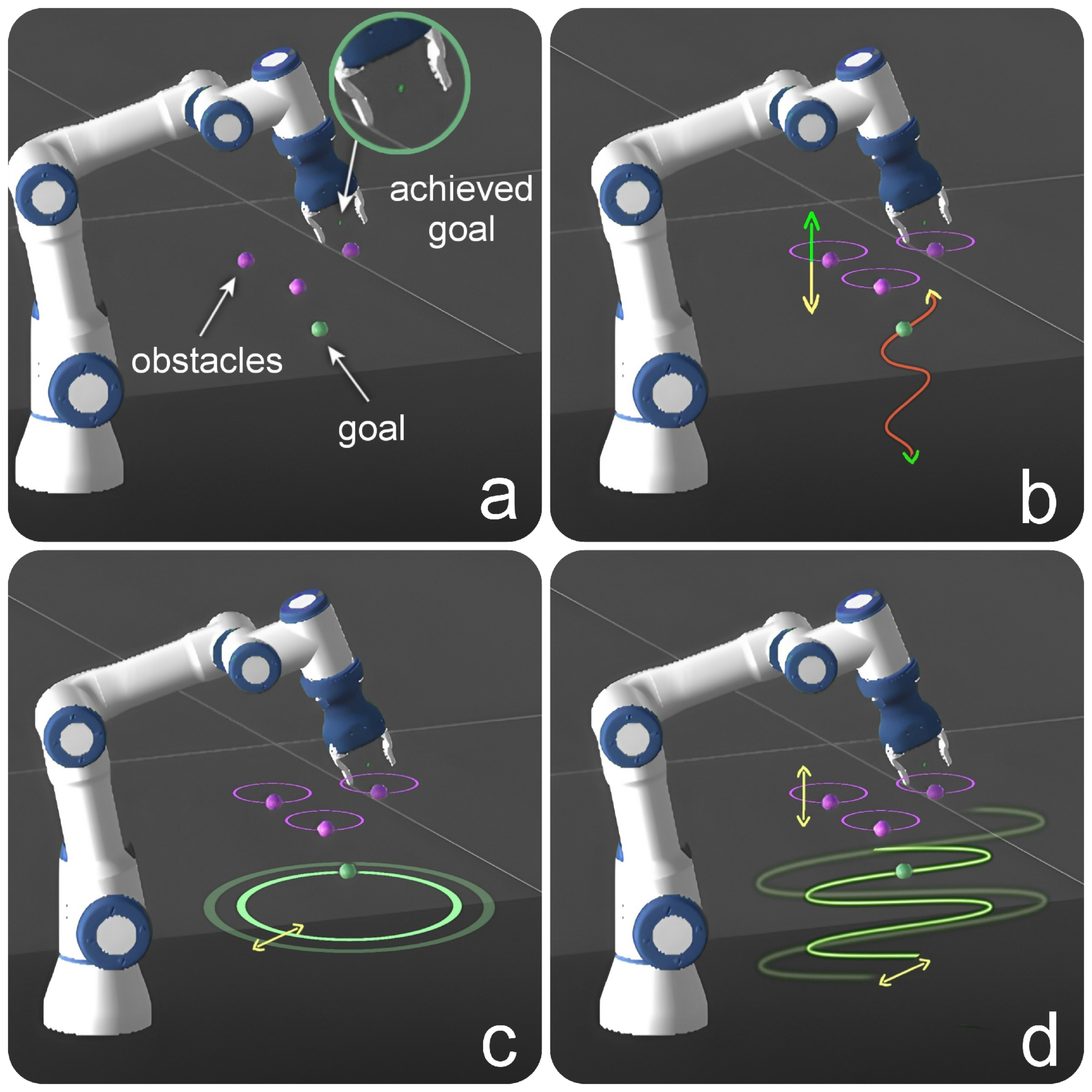}
\caption{Robot environment for testing CONTHER to follow Complex Trajectory with Obstacles Task with Goal, Obstacles, Achieved Goal and experimental trajectories: Path following settings (a), Sinusoid (b), Circle (c), Spiral (d).}\label{fig:shapes_pic}

   \vspace{-0.4cm}
\end{figure}

%In goal-oriented tasks, the agent's objective is to achieve a specified goal state. In such scenarios, the actual reward function is typically set to zero for transitions that do not lead to the goal state, providing a positive reward only in that instance. To speed up model convergence, there are techniques for specialized sampling of buffer transitions. 
In goal-oriented tasks, the agent receives a positive reward only upon achieving the goal, with zero reward for all other transitions. To accelerate convergence in such sparse reward settings, techniques for specialized sampling of replay buffer transitions have been proposed.
For example, Prioritized Experience Replay (PER), proposed by Schaul et al.\cite{per}, allows the agent to emphasize transitions in a buffer with high prediction error. It is also possible to generate intermediate goals for the agent based on the achieved states to simplify the task\cite{subgoals}.

Another solution to the sparse reward function issue in goal-oriented tasks is Hindsight Experience Replay (HER), which was introduced by Andrychowicz et al. \cite{her}. 
%The training dataset for value networks can be expanded using HER in conjunction with an off-policy RL algorithm.
In HER, states visited during training serve as alternative target states and can lead to a high reward during training, even the real target state has not been reached. HER has proven its applicability to many robotic manipulation tasks \cite{liu2022goalconditionedreinforcementlearningproblems}.

%In numerous real-world scenarios, an agent confronts circumstances in which the current state of the environment is inadequate in providing all the requisite information to reach an optimal decision. In such instances, it becomes imperative to take into account the sequence of prior actions and states, i.e. the context.
In many real-world scenarios, the current environmental state does not provide all information necessary for optimal decision-making. In such cases, the agent must consider the sequence of prior actions and states, i.e. the context, to act effectively.
%, thereby enabling the agent to formulate well-informed decisions.
In recent years, there has been a proliferation of research endeavors aimed at integrating RL methodologies with models that possess the capacity to store and examine sequences of actions and states. 

\begin{figure*}[ht]
  \centering
  \includegraphics[width=0.93\linewidth]{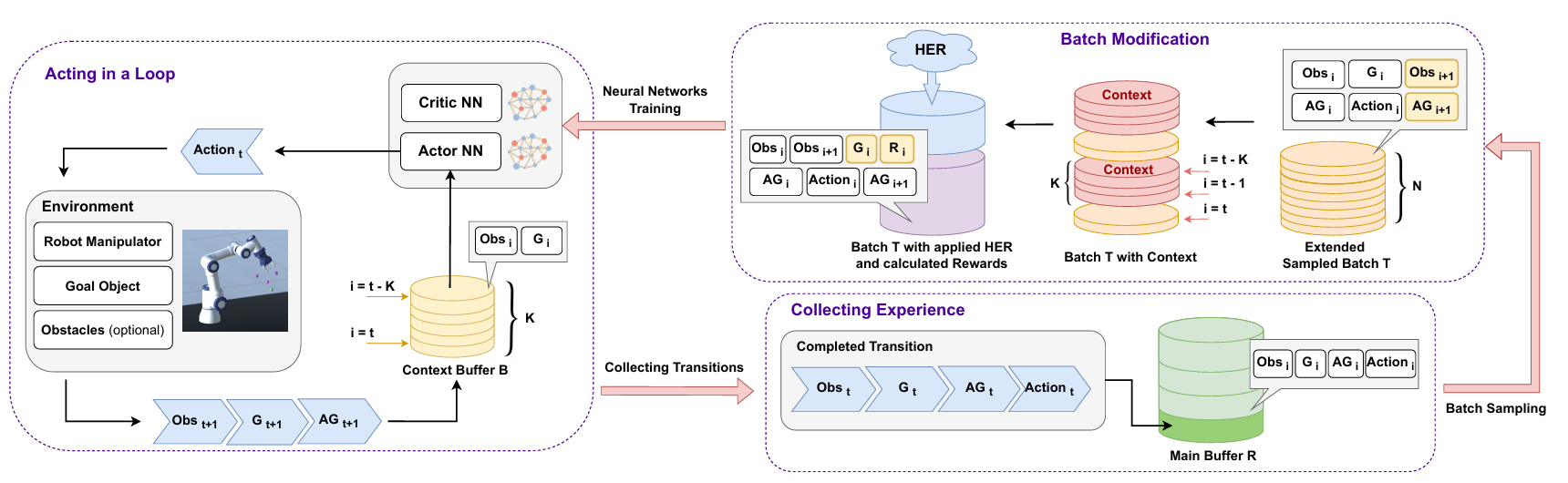}
   \caption{CONTHER Overview: This figure shows the CONTHER architecture, which consists of three main components: Acting in a Loop module, Collecting Experience  module, and Batch Modification module. The Actor in the Acting in a Loop module interacts with the Environment based on the previous context of states (Obs) and goals (G) and passes the collected data to the Collecting Experience module. The Collecting Experience module populates the Main Buffer with vectors of states, goals, achieved goals (AG), and actions, from which batches are periodically sampled. The Batch Modification module adds vectors of future states and achieved goals to a batch of size \textit{N}, and adds \textit{K} previous steps representing the context for each of the time steps. HER (Hindsight Experience Replay) is then applied to a portion of the resulting batch, modifying some of the goal values, after which the reward (R) is calculated for all vectors in the batch. The modified batches are then used to train Actor and Critic neural networks.}\label{fig:mainschema}
   \vspace{-0.4cm}
\end{figure*}

For example, in partially observable environments (POMDP) tasks, the employment of long short-term memory (LSTM) networks has been proposed, enabling the agent to retain information about previous states and actions\cite{pomdp1,podmp2}.
Transformers have demonstrated effectiveness in accounting for long-term dependencies in sequences, particularly in the context of Natural Language Processing (NLP), due to their inherent attention mechanism \cite{vaswani2023attentionneed}. However, it is noteworthy that this ability can be transferred to RL as well, where the consideration of history of actions and states is paramount.

\begin{comment}
The Decision Transformer and Trajectory Transformer were proposed by Chen et al. and Janner et al. \cite{chen2021,janner2021}, which treat RL as a sequential action prediction task based on the history of states, actions, and rewards. In a subsequent study, Action Chunking with Transformers (ACT) algorithm learns a generative model over action sequences \cite{act}. These models demonstrate high performance in the planning and trajectory generation tasks.

Consequently, the question arises: \textit{Why these two approaches, which have been demonstrated to be effective, have not been merged to enhance the model learning process?}

The Transformer Architecture identifies contextual state information for decision-making, while Hindsight Experience Replay (HER) assists the Transformer model in learning from artificial examples of sequential steps of successful behavior leading to high reward. The combination of these approaches has the potential to create a novel approach that not only improves data efficiency but also enhances the agent's ability to generalize across tasks and optimize policies in complex environments. This synergy is of particular importance for achieving faster convergence and better performance in real-world applications, particularly in robotics, where data is often limited and rewards are sparse.
Thus, we built our algorithm, called \textit{CONTHER} (CONText + HER). 
\end{comment}
The Decision Transformer and Trajectory Transformer were proposed by Chen et al. and Janner et al. \cite{chen2021,janner2021}, which treat RL as a sequential action prediction task based on the history of states, actions, and rewards. However, these methods operate offline and rely on pre-collected demonstration datasets, limiting their applicability in online robotic settings where expert data is scarce. In a subsequent study, Action Chunking with Transformers (ACT) \cite{act} learns a generative model over action sequences, but still requires demonstration data.

Consequently, the question arises: \textit{Why have these effective Transformer-based approaches not been combined with HER?}

The Transformer architecture encodes contextual state information for decision-making, while HER assists in learning from artificially successful trajectories. \textit{CONTHER} (CONText + HER) merges both by integrating a Transformer encoder with HER within an online actor-critic algorithm, operating directly on joint velocities and requiring no expert demonstrations.
To our knowledge, this is the first work to combine sequential goal-conditioned relabeling with explicit motion history for robotic manipulation. 
%This synergy achieves faster convergence and better performance in sparse-reward environments, as we demonstrate experimentally. 
%Thus, we built our algorithm, called \textit{CONTHER} (CONText + HER).

The main contributions of the presented work
can be summarized as follows:
\begin{itemize}

\item An RL algorithm \textit{CONTHER} is proposed, which integrates a Transformer architecture for contextual learning with a Hindsight Experience Replay (HER) technique to enable context analysis and artificial sampling of successful trajectories within a buffer.

\item The \textit{CONTHER} outperforms the other considered baseline algorithms by an average of 38.46\%, surpassing the most effective algorithm by 28.21\%, in the task of reaching a point by a robot manipulator. The \textit{CONTHER} learning methodology has also been evaluated with complex dynamic path following and obstacle avoidance tasks. 

\end{itemize}

\section{CONTHER Approach} \label{conther}{

\subsection{Motivation for Contextual Fusion}{

HER addresses reward sparsity by relabeling goals in past trajectories \cite{her}. However, its standard formulation operates on individual transitions $(s_t, g_t, a_t, s_{t+1})$, treating each step independently. 
%This atemporal perspective presents a fundamental limitation for robot control: in joint-space manipulation, the current state $s_t$ contains no information about the dynamic of motion. 
This atemporal perspective presents a fundamental limitation for robot control: in joint-space manipulation, the current state \(s_t\) captures only instantaneous configuration, offering no information about motion dynamics. While augmenting observations with finite differences between consecutive states could provide velocity estimates, such local approximations remain insufficient for capturing higher-order motion characteristics including acceleration, smoothness, and coordinated multi-joint patterns that unfold over longer timescales. Determining whether the end-effector is moving toward or away from the goal with consistent intent requires access to a broader temporal context \(H_{t-1}\), which encodes the manipulator's motion history, essential for generating coherent trajectories.
Formally, the optimal policy is not $\pi(a_t | s_t, g)$ but rather $\pi(a_t | s_t, g_t, H_{t-1})$, where $H_{t-1} = \{s_{t-k}, g_{t-k}\}_{k=1}^K$ encodes the manipulator's motion dynamics essential for generating smooth and physically plausible motions. \textit{CONTHER} approximates this function by explicitly modeling $H_{t-1}$ through a Transformer architecture, transforming an atemporal HER-based learner into a temporally-aware decision-making system. The complete algorithm, built upon TD3 \cite{td3} for stability in continuous control, is detailed below and illustrated in Fig. \ref{fig:mainschema}.}

}

%\vspace{-0.5cm}
\subsection{CONTHER Buffer Sampling}{

%During training, completed episodes are sent to the Main Buffer (named $\mathbb{R}$ in the CONTHER description) for later sampling and model updates. A generalized scheme of the Main Buffer can be seen in Fig. \ref{fig:buffer} (a). It stores information about Observations (Obs), Achieved Goals (AG), Episode Goals (G), Agent Actions (Actions), as well as Next Observations (N-Obs) and Next Achieved Goals (N-AG) obtained from Obs and AG by simple shift. Reward information is also stored in the Main Buffer, but these values are not sampled during playback from the environment,  but are directly calculated within the buffer based on the values of goals set and achieved, the latter of which can be modified using the Hindsight Experience Replay technique.

During training, completed episodes are sent to the Main Buffer (named $\mathbb{R}$ in the \textit{CONTHER} description) for later sampling and model updates. A generalized scheme of the Main Buffer can be seen in Fig. \ref{fig:buffer}. It stores information about Observations, Achieved Goals, Episode Goals, Agent Actions, as well as Next Observations and Next Achieved Goals obtained from Observations and Achieved Goals by simple shift. Reward information is also stored in the Main Buffer, but these values are not sampled during playback from the environment, instead they are calculated directly within the Buffer based on the values of goals set and achieved, the latter of which can be modified using the HER technique. 

%The Main Buffer contains episodes, each of which consists of several steps. Some of the episodes (highlighted in cyan), in each of which specific steps are selected (highlighted in magenta), are selected to sample the training batch. For each such selected step, K previous steps are selected from the buffer to preserve the context (highlighted in yellow). Then some sets of K+1 steps are selected for HER to be applied (shown schematically with green arrows), while the rest remain unchanged.  

\begin{figure}[]
  \centering
   %\subfigure[Main Buffer.]{\includegraphics[width=0.44\linewidth]{UFFER_.png}}
   %\subfigure[Sampled Batch.]{\includegraphics[width=0.44\linewidth]{uffer_a_.png}}
   \includegraphics[width=0.99\linewidth]{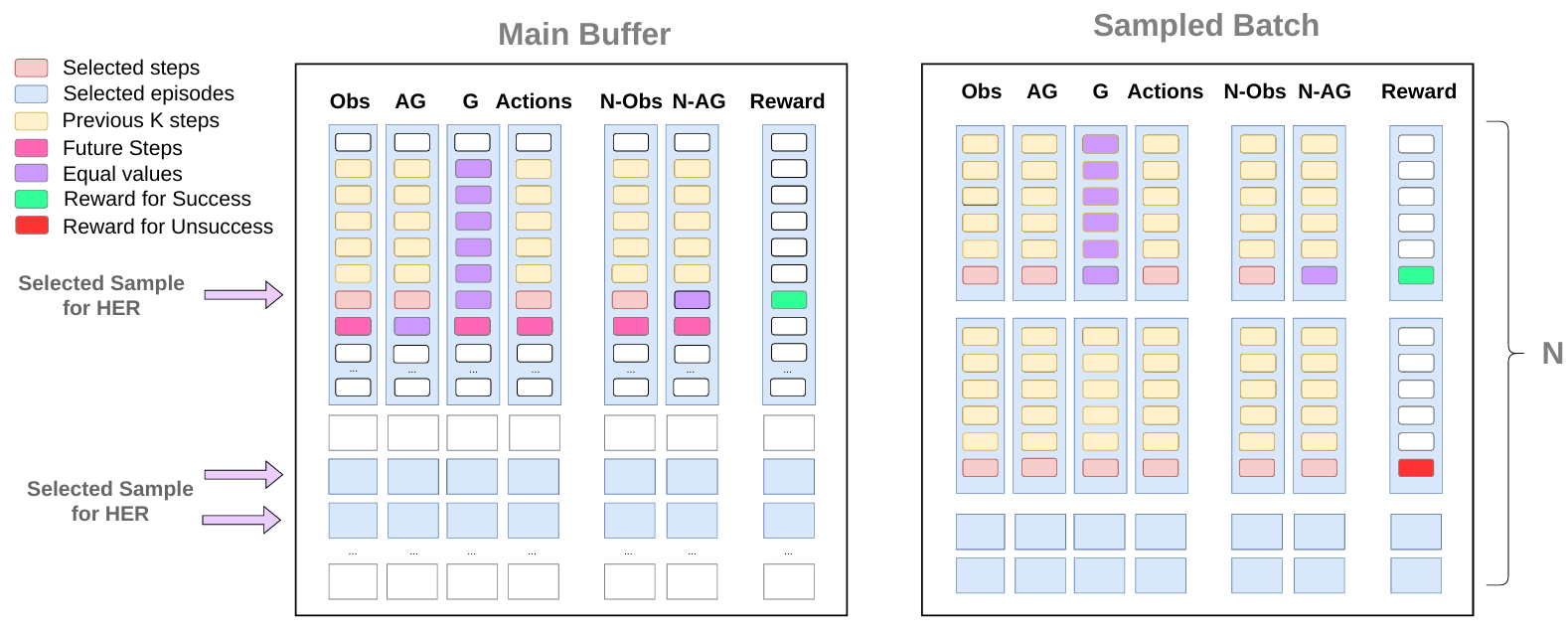}
    
  \caption{Buffer modification for Sampling during Training. Stored vectors: Observations (Obs), Achieved Goals (AG), Episode Goals (G), Agent Actions (Actions), Next Observations (N-Obs), Next Achieved Goals (N-AG), Reward.}\label{fig:buffer}
   
   \vspace{-0.5cm}

\end{figure}

The Main Buffer contains episodes, each of which consists of several steps. Some of the episodes, each with specific steps chosen, are selected to sample the training batch. For each chosen step, \textit{K} previous steps are selected from the buffer to preserve the context. Then some sets of \textit{K}+1 steps are selected for HER to be applied, while the rest remain unchanged.

For the \textit{K}+1 samples selected to apply HER, we use the subsequent step \textit{K}+2 for relabeling. Within a block of \textit{K}+2 consecutive transitions, we apply HER by setting the Achieved Goal of the last transition (step \textit{K}+2) as the Future-AG. This value is then substituted for all Goal vectors of the first \textit{K}+1 transitions and for the Next Achieved Goal at step \textit{K}+1. Consequently, the entire block of \textit{K}+1 transitions is relabeled as a coherent successful trajectory.

%For a block of \textit{K}+1 consecutive transitions selected for HER, we take an achieved goal Future-AG from a later \textit{K}+1 time step and replace all \textit{K} Goal vectors and the Next Achieved Goal of the \textit{K}-step transition with the same Future-AG value. This ensures that the entire context block is relabeled consistently as a successful trajectory.
%Then, the values of all \textit{K}+1 Goals and Next Achieved Goal in the last step are manually substituted with an equal value of Achieved Goal from the future. Thus, in the transformed sample, the agent artificially achieves the Goal given the context from \textit{K} previous steps. This allows the agent in this case to assign Reward for Success. 

Fig. \ref{fig:buffer} shows the final view of a Sampled Batch of \textit{N} elements, each containing \textit{K}+1 consecutive steps. Some of these episodes will have Reward for Unsuccess and some will have Reward for Success, which can be achieved with or without using HER.

\begin{algorithm}[ht]
\caption{CONTHER(+TD3) Algorithm}
\label{alg:conther}
\begin{algorithmic}[1]

\Require Critic networks $Q_1(\theta_1^Q), Q_2(\theta_2^Q)$, actor $\mu(\theta^\mu)$
\Require Target nets $Q_1',Q_2',\mu'$ with soft update $\tau$
\Require Buffer $R$, context size $K$, batch $N$, HER rate, noise $\mathcal{N}$

\For{episode $=1$ to $M$}
    \State Get $s_1$, goal $g_1$, achieved goal $ag_1$
    \State Init context $B = \{(s_i,g_i)\}_{i=1}^K$
    
    \For{$t = 1$ to $T$}
        \State $a_t = \mu(B;\theta^\mu) + \mathcal{N}_t$
        \State Store $(s_t,g_t,a_t,ag_t)$ in $R$
        \State Execute $a_t$, observe $s_{t+1},g_{t+1},ag_{t+1}$
        \State Update $B$ with $(s_{t+1},g_{t+1})$
    \EndFor
    
    \For{$t = 1$ to $S$}
        \State Sample $N$ transitions $T'$ from $R$
        \State Get $T'$ with next states $(s_{i+1},ag_{i+1})$
        \State $m = [i-K;i]$ \Comment{get $K$ previous steps}
        \State Extend $T'$ with $(s_m,g_m,a_m,ag_m,s_{m+1},ag_{m+1})$
        \State Apply HER to $N'<N$ samples: $g_m \leftarrow g_m'$
        \State $G_m = [g_m \text{ or } g_m']$
        \State \brokenline{% 
        Compute $r_i$ for $(s_m,G_m,a_m,ag_m,s_{m+1},ag_{m+1})$}
        
        \State \brokenline{% 
        $y_i = 
        r_i + \gamma \min_j Q_j'(s_{m+1},G_{m+1},\mu'(s_{m+1},G_{m+1}))$}
        
        \State \brokenline{% 
        Update critics: 
                $L_j = \frac{1}{N}\sum_i[y_i - Q_j(s_m,G_m,a_i;\theta_j^Q)]^2$}
        
        \State 
        \If{$t \bmod w = 0$}
           \State \brokenline{%
           Update actor: \begin{equation*}
            L = \frac{1}{N}\sum_i[\alpha\mu(s_m,G_m;\theta^\mu)^2 -\end{equation*} \begin{equation*} - \min_j Q_j(s_m,G_m,\mu(s_m,G_m;\theta^\mu))]
            \end{equation*}}
        \EndIf
        
    \EndFor
    
    \State $\theta^{Q_j'} \gets \tau\theta^{Q_j} + (1-\tau)\theta^{Q_j'}$
    \State $\theta^{\mu'} \gets \tau\theta^{\mu} + (1-\tau)\theta^{\mu'}$
\EndFor

\end{algorithmic}
\end{algorithm}

\subsection{Models Architecture} \label{models}

 Generalized schematics of the Actor and Critic models are shown in Fig. \ref{fig:act_crit}. They consist of a Transformer Blocks as well as Fully Connected Layers. Each of the Transformer Blocks was implemented based on the TrXL-I architecture proposed by Parisotto et. al. \cite{parisotto} to increase the stability of Transformers in RL. The Fully Connected Layers have a depth of three.

The architectural distinction between \textit{CONTHER-v.0} and  \textit{CONTHER-v.1} investigates information sufficiency in contextual policy learning. In both variants, the Actor receives \textit{K}+1 consecutive observation-goal pairs processed through a Transformer Block. We hypothesize that while the Transformer captures long-range dependencies, its compressive representations may dilute precise current state information critical for accurate control. To evaluate this hypothesis, \textit{CONTHER-v.1} introduces a residual-style connection that concatenates the Transformer output with the raw last-step input vector, thereby preserving both temporal context and current state detail before passing through Fully Connected Layers. \textit{CONTHER-v.0} omits this connection, relying solely on the transformed context. This architectural ablation assesses whether explicit current state information remains necessary when sequential context is available. Experimental results are presented in Section \ref{exps}-B.

\begin{figure}[ht]
  \centering
  \subfigure[Actor Model Architecture.]{ \includegraphics[width=0.89\linewidth]{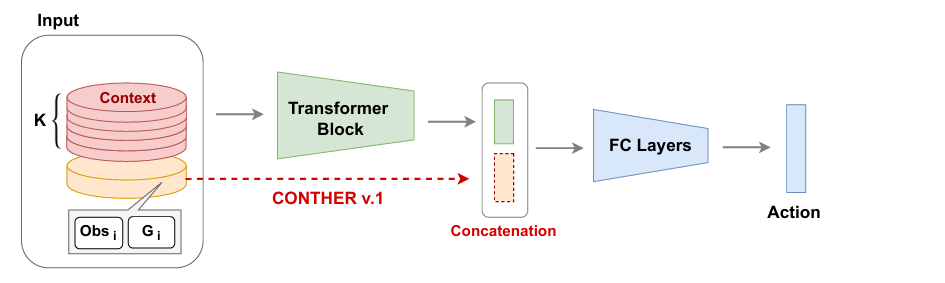}}
   \subfigure[Critic Model Architecture.]{\includegraphics[width=0.89\linewidth]{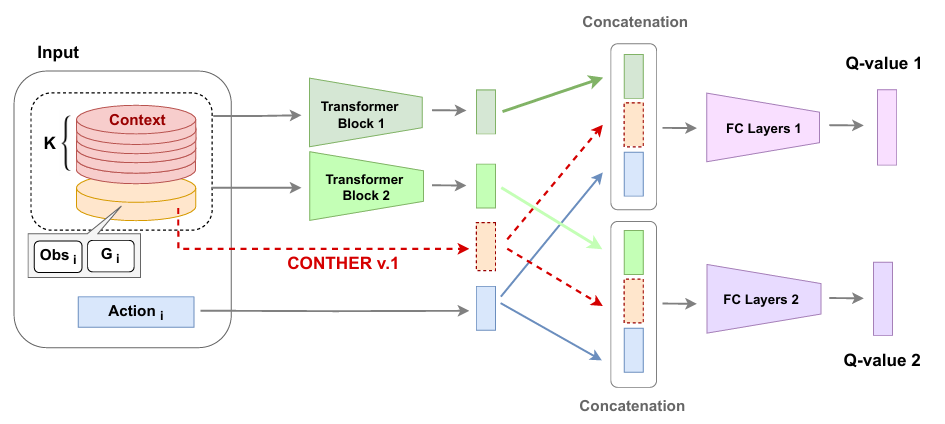}}
\caption{Actor and Critic networks architecture for CONTHER-v.0 and CONTHER-v.1. The connection highlighted in red is present only in CONTHER-v.1.}\label{fig:act_crit}
   \vspace{-0.3cm}
\end{figure}
%\vspace{0.5cm}

%In the Critic model, there are two parallel architectures, estimating \textit{Q}-values.

In the algorithm described earlier, two different Critic's models were presented for a clearer explanation, but in our implementation two parallel architectures are merged into one, estimating two \textit{Q}-values. The input to the models is also a set of \textit{K}+1 sequential vectors, \textit{K} of which represent the prior context for the last of the steps, and the Agent's Actions vector for the last step. The context vector is transformed by two parallel Transformer Blocks, then in \textit{CONTHER-v.1} the output of each of them is concatenated with the last context vector and the Actions vector. The resulting vectors are passed through two decoupled sets of Fully Connected Layers, resulting in the estimation of two Q-values. In \textit{CONTHER-v.0}, the architecture is the same, except for the concatenation with the last step context vector after Transformer Blocks.

\section{Experiments}\label{exps}

This study investigates whether explicit sequential context improves HER based policy learning in goal conditioned manipulation. Addressing this question requires isolating the effect of context from confounding factors such as sensor noise and mechanical variability. Simulation provides the controlled environment necessary for this isolation, enabling precise measurement of contextual influence across multiple trajectories and random seeds. Furthermore, the dynamic tasks employed, including moving goals, oscillating obstacles, and coordinated multi axis motions, require exact reproducibility that is inherently infeasible with physical hardware. Simulation is therefore the appropriate methodological tool for validating the core contribution prior to real world deployment.
 
\subsection{Environment Description.} 
% ДОБАВИТЬ ОБСЕРВЕЙШНС, перенести сюда описание ревордов, KKKKKKK
The environment is based on the Unity engine \cite{unity}. The Unity environment contains a robot model (Universal Robots UR3), a Robotiq Hand-E gripper, and a target object to be reached. The robot is controlled by the speed of the 6 joints in different  pre-selected proportions for each joint. 

The geometric space of possible locations of the target object is located around the robot and resembles the shape of a torus slightly raised from the surface on which the robot is mounted.

The Observation vector includes information about the position and rotation of the joints in the global coordinate system, the angle of their own rotation, as well as the target object and end-effector vectors in the robot coordinate system. The point centered between the gripper's fingers is the point that should coincide as closely as possible with the location of the target object (Fig. \ref{fig:shapes_pic}(a)). The neural networks reside in a Python environment and communicate with Unity via TCP. 

The context length \textit{K}$=$6 was chosen via a hyperparameter sweep on the point-reaching task. \textit{K}$=$4 resulted in unstable learning (success rate$<$50\%), while larger values (8,10) did not improve performance.

\subsection{Comparative Experiment on the Reaching Point Task.}
Since the Point Reaching task is quite general to other more complex tasks, it was chosen to experiment and compare the \textit{CONTHER} algorithm with others.

\textbf{{Algorithms considered:}} The experiment compared 5 different approaches, referred to as \textit{CONTHER-v.0}, \textit{CONTHER-v.1}, \textit{TD3+Context}, \textit{TD3+HER} and \textit{TD3}.  \textit{CONTHER-v.0} and \textit{CONTHER-v.1} were introduced and described in Section \ref{conther}. To isolate the individual contributions of the context mechanism and the HER strategy, we designed our baselines as incremental modifications of the TD3 algorithm. This allows for a direct attribution of performance gains to the specific components proposed in this paper. \textit{TD3}, \textit{TD3+HER} imply the Actor and Critic architectures described in Section \ref{conther}-B, but the input is a vector without context, i.e. only one step of the Agent is affected. \textit{TD3+HER} differs from \textit{TD3} by using the HER technique to artificially sample successful trajectories in the Buffer.  \textit{TD3+Context} differs from \textit{CONTHER-v.1} by not using the HER technique. 
%Thus, all possible modifications of the learning algorithms have been constructed to fully describe and prove the validity of the \textit{CONTHER-v.1} algorithm. 

The reward was chosen to be sparse, $r=0$ when the target point area is reached by a distance $d \leq d_{threshold}$ and $r=-1$ when $d > d_{threshold}$. In this experiment, $d_{threshold} = 0.1$ was determined empirically from $[0.05,0.1,0.15,0.2]$. This value balanced the need for sufficient reward density during training against the requirement for precise final positioning, as smaller thresholds hindered learning while larger ones led to coarse policy convergence.

The ratio between HER replays and regular replays in the future strategy was set to 4 as proposed in the original paper\cite{her}. %This configuration balances the introduction of artificially successful trajectories with preservation of the original experience distribution. 
This ratio balances artificially successful trajectories against the original experience distribution.

Training was performed for 25 epochs, each with 50 full episodes and 40 iterations of updating the neural network with the same hyperparameters, model validation was performed after each epoch for 10 episodes, and the results are presented in Fig. \ref{fig:reaching}(a,b,d,e).  

\begin{figure*}[ht]
  \centering
  \subfigure[Actor Loss.]{ \includegraphics[width=0.27\linewidth]{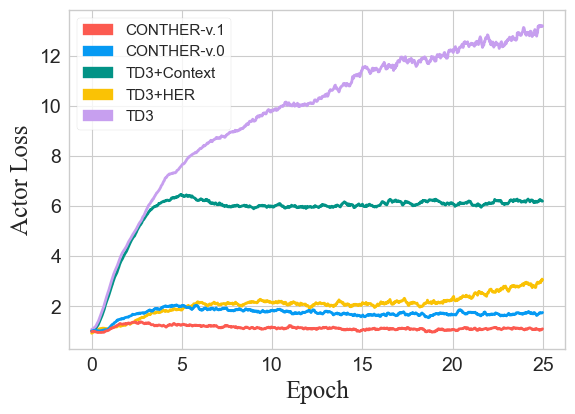}}
   \subfigure[Critic Loss.]{\includegraphics[width=0.27\linewidth]{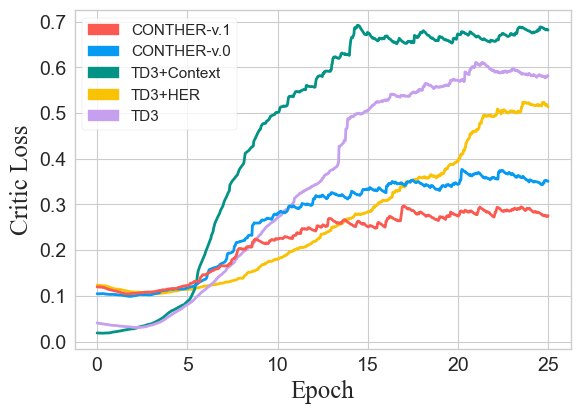}}
   \hspace{3em} 
   \subfigure[Mean Reward.]{ \includegraphics[width=0.27\linewidth]{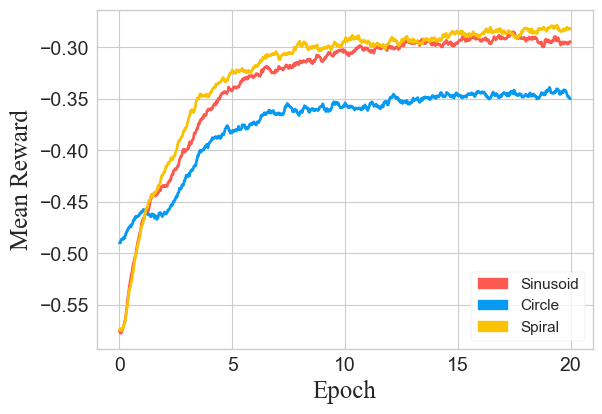}}

   \subfigure[Mean Reward.]{\includegraphics[width=0.27\linewidth]{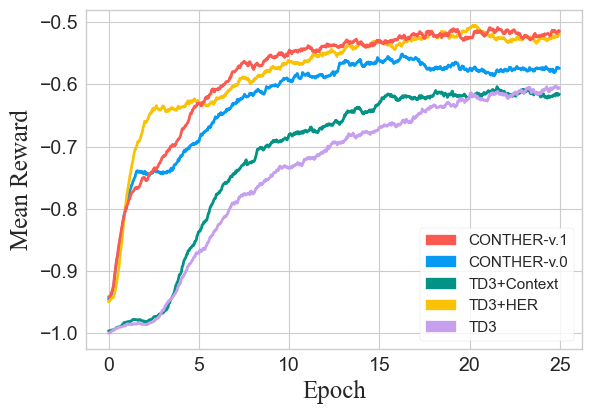}}
    \subfigure[Mean Success Rate.]{\includegraphics[width=0.27\linewidth]{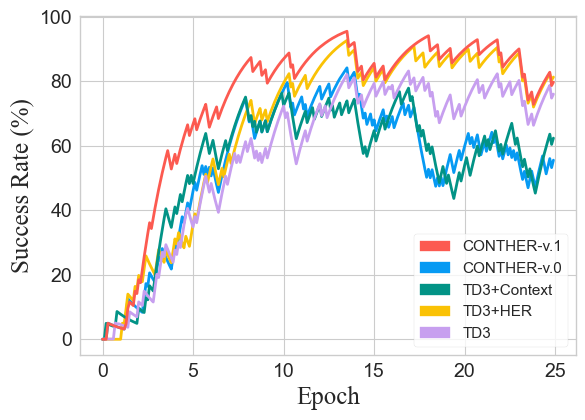}}
    \hspace{3em} 
    \subfigure[Mean Success Rate.]
   {\includegraphics[width=0.27\linewidth]{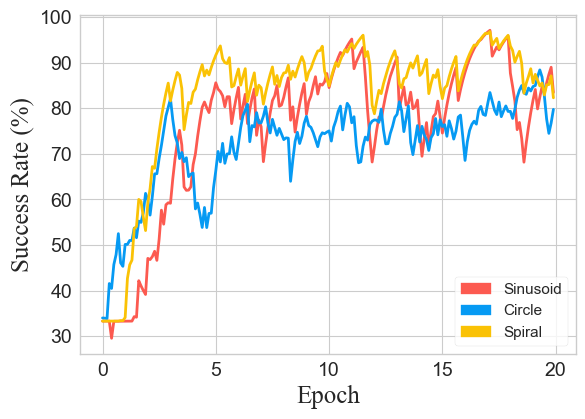}}
   \caption{Algorithm benchmarking for the Reaching Point Task (a,b,d,e) ; CONTHER-v.1 training results for Complex Trajectory Following with Obstacles Task for three types of trajectories: Sinusoid, Circle, Spiral (c,f). Mean Success Rate means Percentage of transitions that have a success reward during validation.}\label{fig:reaching}
   
   \vspace{-0.2cm}
\end{figure*}

Fig. \ref{fig:reaching}(a) and Fig. \ref{fig:reaching}(b) illustrate the Actor and Critic losses behavior. As can be seen, the losses converge to the final value using the context-aware mechanism, namely \textit{CONTHER-v.0}, \textit{CONTHER-v.1} and \textit{TD3+Context}, which illustrates the validity of its embedding. The best metrics were shown by \textit{CONTHER-v.0} and \textit{CONTHER-v.1}, the latter of which outperformed the other due to the presence of additional link within the architecture described in \ref{models}.

Fig. \ref{fig:reaching}(d) and Fig. \ref{fig:reaching}(e) illustrate the Mean Reward values during training and the Success Rate of the agent during validation. As can be seen, the leading position is taken by \textit{CONTHER-v.1} and the \textit{TD3+HER} baseline. Notably, as seen from the Success Rate values, \textit{CONTHER-v.1} outperforms all other approaches both in terms of the rate of achieving a high success rate and in terms of remaining the highest throughout the experiment.

This experiment demonstrates the most stable behavior of the \textit{CONTHER-v.1} model during training, which not only leads to a fast convergence of the Losses values, the lowest of all, but also to a high Success Rate, outperforming the other algorithms in the last stages of training (5 last epochs) by an average of 38.46\%, and the most successful baseline \textit{TD3+HER} by 28.21\%. This defines \textit{CONTHER-v.1} as the most successful architecture of all considered.

\subsection{Complex Trajectory Following with Obstacles.}

Given that \textit{CONTHER} involves the consideration of context in decision-making processes, a series of experiments were designed to test the ability to predict motion trajectory based on prior context with complicating conditions in the form of obstacles. 

Obstacles were added to the environment. These obstacles were located above the goal and with some randomness in range (Fig. \ref{fig:shapes_pic}(a)). This arrangement was chosen because in previous experiments the robot reached the goal by moving towards it from the top downwards or at an angle, so the robot is more likely to pass through the area containing these obstacles. It should also be noted that, due to the increased complexity of the task, the amplitudes of the robot joint movements corresponding to the maximum Action value were halved compared to the experiment in Section \ref{exps}-B.

However, when applying \textit{CONTHER} in the presence of additional obstacles, the algorithm requires a minor modification. Objects to be avoided can be treated as goals \textit{G-} that should \textbf{not} be achieved, in contrast to the positive goals \textit{G+}. In this representation, the buffer modification for generating artificially successful trajectories, described in Section \ref{conther}-A, is applied as follows. We relabel the positive goals \textit{G+} with future achieved goals, as was done before. For the negative goals \textit{G-}, we do not leave them fixed; instead, we recompute their positions so that they maintain the same relative offset to the modified \textit{G+} as they had to the original \textit{G+}. In other words, the obstacles are translated by the same displacement vector as the positive goal. 

\begin{figure}[h]
  \centering
  \includegraphics[width=0.9\linewidth]{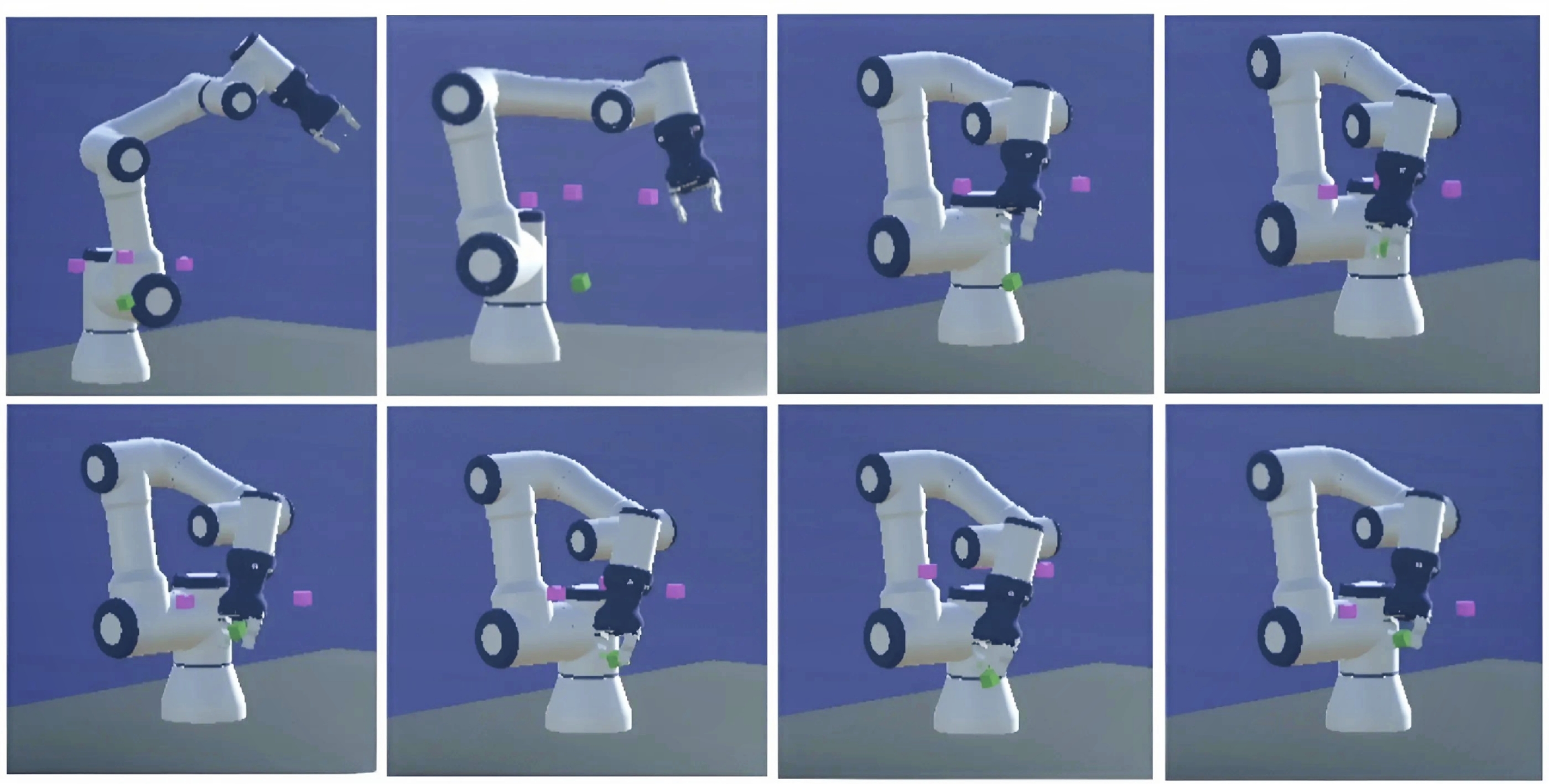}
\caption{Example of the Task of Following Complex Sinusoid Trajectory with Obstacles (from top left to bottom right).}\label{fig:traj}
   \vspace{-0.2cm}
\end{figure}

Three complex trajectories were chosen, as shown in Fig. \ref{fig:shapes_pic}. The first trajectory (Fig. \ref{fig:shapes_pic}(b), Fig. \ref{fig:traj}) is a vertical sinusoid whose amplitude changes each episode, with obstacles moving in the opposite vertical direction and circling around their axes. This requires the agent to continuously modulate Z‑axis velocity to track periodic motion under dynamic obstacle interference. The second trajectory (Fig. \ref{fig:shapes_pic}(c)) is a horizontal circle whose radius changes after each full cycle, while obstacles repeat the target's motion with additional circular rotations. This tests planar tracking with varying curvature, a key skill for coordinated multi‑axis control. The third trajectory (Fig. \ref{fig:shapes_pic}(d)) is a spiral combining forward and backward motion, with amplitude varying from lower to upper bounds per traversal; obstacles move with the target while oscillating vertically and rotating. This demands simultaneous coordination across all six joints, testing the agent's ability to manage coupled multi‑axis dynamics under continuous perturbation. In all three scenarios, anticipatory planning and context‑driven path adjustment are further evaluated through moving obstacles, which represent essential capabilities for dynamic environments.

In each of the three types of trajectories, the initial positions of the goal and obstacles in the horizontal plane are random. To capture the context, the update interval for all point positions is set to half the length of the context buffer \textit{K}. The hyperparameters of the trajectories in each of the three cases were chosen to complicate the learning tasks.

The reward in these tasks is composed of two terms related to reaching the goal and bypassing obstacles: 
\begin{itemize}

\item  $r_{goal} =0$ when the target point area is reached by a distance $d_{goal} \leq d_{g - threshold}$ and $r=-1$ when $d_{goal} > d_{g - threshold}$. In this experiment, $d_{g - threshold} = 0.09$ .

\item $r_{obst} =0$ when the distances to each of the three obstacles are $d_{obst} \geq d_{obst - threshold}$ and $r=-1$ otherwise. In this experiment, $d_{obst - threshold} = 0.05$.  

\item The final formula for reward is: $ r = 0.5 \times( r_{goal} + r_{obst}).$ 
    
\end{itemize}

Training was performed for 20 epochs, each with 50 full episodes and 40 iterations of updating the neural network.
The ensuing training outcomes are depicted in Fig. \ref{fig:reaching}(c,f). As can be seen, the movement involving only the horizontal plane seemed to be more difficult for the agent.

However, in all three tasks focusing on the necessary analysis of the prior context in different formulations, the \textit{CONTHER-v.1} algorithm showed convergence and high success rate during validation,
%by an average of 78.54\%,
proving its robustness in dealing with such complex tasks involving obstacle avoidance and following a dynamic, frequently changing trajectory.

\section{Conclusions and Future Work.}

In this work, we presented \textit{CONTHER}, an RL algorithm that integrates a Transformer architecture for contextual learning with HER to generate successful trajectories in the replay buffer, eliminating the need for expert demonstrations. In a point-reaching task with a robotic manipulator, \textit{CONTHER} outperformed baseline algorithms, achieving an average improvement of 38.46\% across all baselines and a 28.21\% advantage over the strongest baseline. Its effectiveness was further demonstrated in complex scenarios requiring dynamic goal following and obstacle avoidance.

Future work will extend the observation space to include varied scenes, objects, and manipulator models. Because \textit{CONTHER} operates directly on joint velocities, transferring it to a new robot only requires changes to the kinematic model and task-specific hyperparameters. Despite being evaluated in simulation, \textit{CONTHER}’s modular design and joint-velocity control enable direct deployment on a physical UR3  in planned future experiments. Recent work on agentic RL \cite{agenticrl} shows that automated reward generation and policy diagnosis reduce manual tuning, and integrating this into \textit{CONTHER} could further ease reward engineering. Furthermore, the proposed principle of contextual replay can be applied to other algorithms and tasks, reducing reliance on expert demonstrations and improving training efficiency for goal-oriented robotic tasks.

\section*{Acknowledgements} 
Research reported in this publication was financially supported by the RSF grant No. 24-41-02039.

\bibliographystyle{IEEEtran}
\bibliography{root}
%\balance{}
\end{document}